\documentclass[conference]{IEEEtran}
\usepackage{times}

\IEEEoverridecommandlockouts                              
\usepackage[numbers]{natbib}
\usepackage{multicol}

\usepackage{xcolor}

\definecolor{MyDarkBlue}{rgb}{0,0.08,1}
\definecolor{MyDarkGreen}{rgb}{0.02,0.6,0.02}
\definecolor{MyDarkRed}{rgb}{0.8,0.02,0.02}
\definecolor{MyDarkOrange}{rgb}{0.40,0.2,0.02}
\definecolor{MyPurple}{rgb}{111,0,255}
\definecolor{MyRed}{rgb}{1.0,0.0,0.0}
\definecolor{MyGold}{rgb}{0.75,0.6,0.12}
\definecolor{MyDarkgray}{rgb}{0.66, 0.66, 0.66}

\usepackage{graphics} 
\usepackage{epsfig} 
\usepackage{mathptmx} 
\usepackage{times} 
\usepackage{amsmath} 
\usepackage{amssymb}  
\usepackage{color}
\usepackage{gensymb}
\usepackage{float}
\usepackage{amsmath}
\usepackage[pagebackref=true,breaklinks=true,letterpaper=true,colorlinks,bookmarks=false]{hyperref}
\newcommand{\myparagraph}[1]{\vspace{0.05in}\noindent\textbf{#1}}

\begin{document}

\title{\LARGE \bf
	Cable Manipulation with a Tactile-Reactive Gripper 
}


\author{
	\authorblockN{Yu She$^{*}$, Shaoxiong Wang$^{*}$, Siyuan Dong$^{*}$, Neha Sunil, Alberto Rodriguez and Edward Adelson}
	\authorblockA{
		Massachusetts Institute of Technology\\
		{\tt\small <yushe,wang\_sx,sydong,nsunil,albertor>@mit.edu, adelson@csail.mit.edu}} 
	\href{http://gelsight.csail.mit.edu/cable/}{http://gelsight.csail.mit.edu/cable/}
	\thanks{This work was supported by the Amazon Research Awards (ARA), the Toyota Research Institute (TRI), and the Office of Naval Research (ONR) [N00014-18-1-2815]. Neha Sunil is supported by the National Science Foundation Graduate Research Fellowship [NSF-1122374]. This article solely reflects the opinions and conclusions of its authors and not Amazon, Toyota, ONR, or NSF.}
	\thanks{* Authors with equal contribution.}%
}

\maketitle

\vspace{-10pt}
\begin{abstract}
Cables are complex, high dimensional, and dynamic objects. Standard approaches to manipulate them often rely on conservative strategies that involve long series of very slow and incremental deformations, or various mechanical fixtures such as clamps, pins or rings.

We are interested in manipulating freely moving cables, in real time, with a pair of robotic grippers, and with no added mechanical constraints.
The main contribution of this paper is a perception and control framework that moves in that direction, and uses real-time tactile feedback to accomplish the task of following a dangling cable. 
The approach relies on a vision-based tactile sensor, GelSight, that estimates the pose of the cable in the grip, and the friction forces during cable sliding.

%
%
We achieve the behavior by combining two tactile-based controllers: 1) Cable grip controller, where a PD controller combined with a leaky integrator regulates the gripping force to maintain the frictional sliding forces close to a suitable value; and 2) Cable pose controller, where an LQR controller based on a learned linear model of the cable sliding dynamics keeps the cable centered and aligned on the fingertips to prevent the cable from falling from the grip.
%
%
This behavior is possible by a reactive gripper  fitted with GelSight-based high-resolution tactile sensors.
The robot can follow one meter of cable in random configurations within 2-3 hand regrasps, adapting to cables of different materials and thicknesses.
We demonstrate a robot grasping a headphone cable, sliding the fingers to the jack connector, and inserting it.
To the best of our knowledge, this is the first implementation of real-time cable following without the aid of mechanical fixtures.

%
%
%

\end{abstract}

\section{Introduction}

%

Contour following is a dexterous skill which can be guided by tactile servoing. A common type of contour following occurs with deformable linear objects, such as cables. After grasping a cable loosely between the thumb and forefinger, one can slide the fingers to a target position as a robust strategy to regrasp it. For example, when trying to find the plug-end of a loose headphone cable, one may slide along the cable until the plug is felt between the fingers. 


Cable following is challenging because the cable's shape changes dynamically with the sliding motion, and there are unpredictable factors such as kinks, variable friction, and external forces. For this reason, much work on cables (and other deformable linear objects) has utilized mechanical constraints ~\cite{yan2019self, zhu2018dual, nair2017combining}. For example a rope may be placed on a table, so that gravity and friction yield a quasistatic configuration of the cable. A gripper can then adjust the rope configuration, step by step, at a chosen pace. 

\begin{figure}[t]
	\centering
	\includegraphics[width= \linewidth]{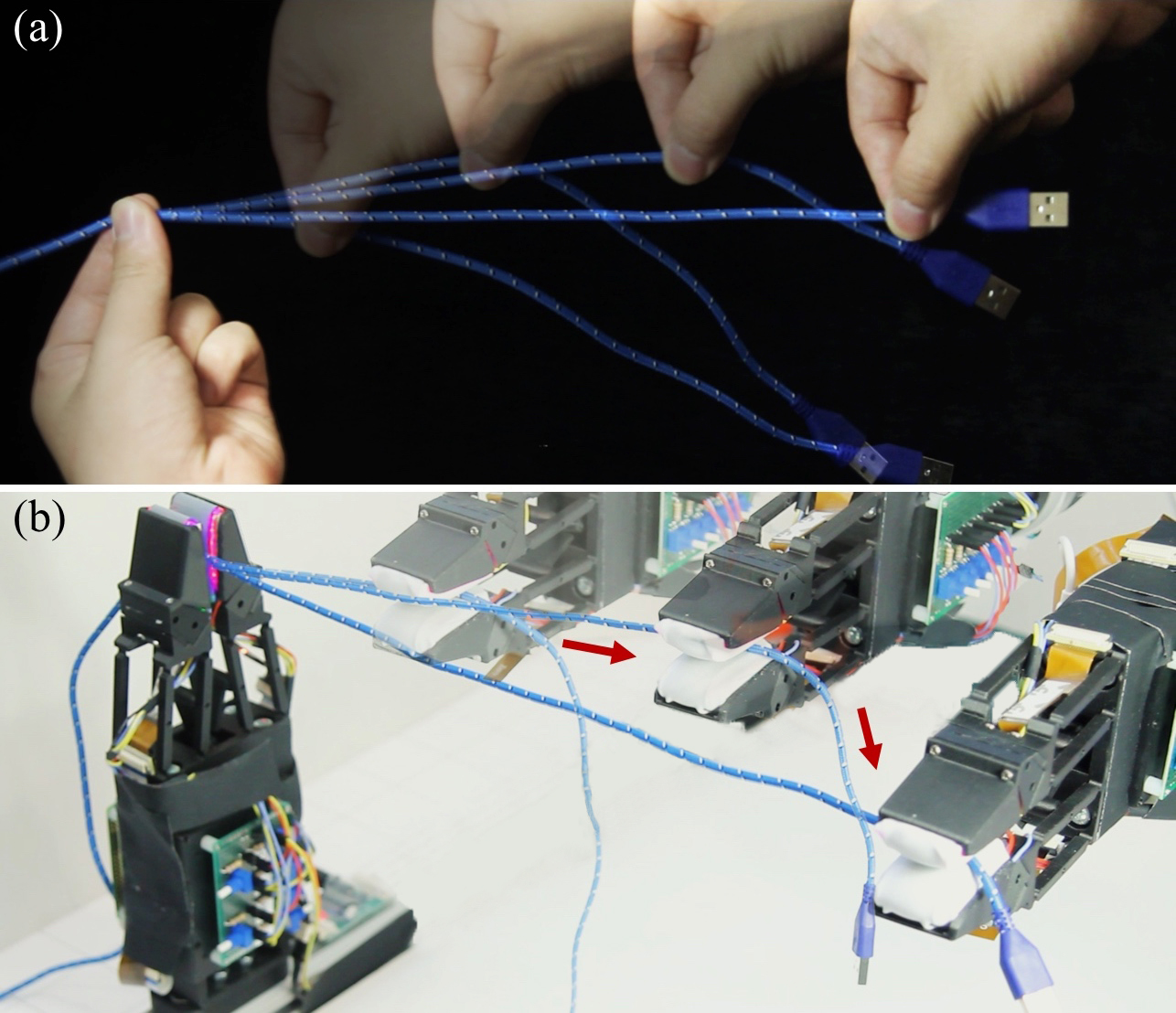}
	\vspace{-10pt}
	\caption{Following a cable with (a) human hands and (b) robotic grippers.}
	\label{fig:introduction}
	\vspace{-10pt}
\end{figure}

Our goal is to manipulate cables in real time, using a pair of grippers, with no added mechanical constraints. The cables are free to wiggle, swing, or twist, and our grippers must rapidly react using tactile feedback. In particular, we look at the task of picking one end of a cable with a gripper and following it to the other end with a second gripper, as shown in Fig.~\ref{fig:introduction}. 

We designed a novel gripper that is lightweight and fast reacting, and equipped it with high resolution tactile sensors. This novel hardware, when paired with appropriate control policies, allows us to perform real-time cable following in free space.

In this paper we do not use vision, relying on tactile sensing alone. While vision can be helpful, we are able to perform the task purely with tactile guidance. Deformable linear objects are easily occluded from view by grippers, by the environment, and often by itself. Tactile perception allows for precise localization once the cable is grasped. Tactile active perception, like when pulling from the two ends of a cable until it is in tension, can also be used to simplify perception such as in the case of a tangled rope.
%





We approach cable following by dividing the desired behavior into two goals: 1) Cable grip control, which monitors the gripping force to maintain friction forces within a useful range; and 2) Cable pose control, which regulates the configuration of the cable to be centered and aligned with the fingers.
To accomplish this task, we build a system with the following modules: 
\begin{itemize}
    \item \textbf{Tactile-reactive gripper.} We design a parallel-jaw gripper with force and position control capabilities (Sec.~\ref{gripper}), fitted with GelSight-based tactile sensors~\cite{gelsight_journal} yielding a 60Hz grip bandwidth control.
    \item \textbf{Tactile perception.} We estimate in real-time the pose of the cable in the gripper, the friction force pulling from the gripper, and the quality of the tactile imprints (Sec.~\ref{sec:perception}).
    \item \textbf{Cable grip controller.} The gripper regulates the gripping force by combining a PD controller and a leaky integrator that modulates the friction force on the cable, and provides grip forces to yield tactile imprints of sufficient quality for perception (Sec.~\ref{sec:control}).
    \item \textbf{Cable pose controller.} The robot controls the cable configuration on the gripper fingers with an LQR controller, based on a learned linear model of the dynamics of the sliding cable (Sec.~\ref{sec:control}).
\end{itemize}



We evaluate the complete system in the task of cable following for various cables, sliding at different velocities, and benchmark against several baseline algorithms. The results in Sec.~\ref{sec:results} show that training the system on a single cable type allows generalization to a range of cables with other physical parameters.
Finally, we demonstrate a robot picking a headphone cable, sliding the fingers until feeling the jack connector, and inserting it, illustrating the potential role of the system in complex active perception and manipulation tasks.


\section{Related Work}
In this section we review work relevant to contour following and cable manipulation. 

\subsection{Contour following} 
Contour following of rigid objects has been widely studied using both visual ~\cite{lange1998viscontour} and tactile perception ~\cite{chen1995edge, tactip}.  These techniques do not directly translate to deformable objects due to dynamic shape changes that are difficult to model, especially in real time.


The most similar contour following work to ours is by Hellman~\textit{et al}.~\cite{hellman2017functional}, who proposed a reinforcement learning approach to close a deformable ziplock bag with feedback from BioTac sensors. The work demonstrated a robot grasping and following the edge of the bag. In contrast to our approach, they use a constant grasping force and discrete slow actions. As a consequence, they achieve a maximum speed of 0.5 cm/s, compared to 6.5 cm/s in our work. 



\subsection{Cable/rope manipulation} 
Manipulating deformable linear objects has attracted attention in the robotics community~\cite{hopcroft1991case} with tasks including tying knots~\cite{morita2003knot, saha2008motion}, surgical suturing~\cite{mayer2008system}, or wire reshaping~\cite{yan2019self, zhu2018dual, nair2017combining}. 

Due to their high dimensional dynamics, manipulating deformable linear objects is usually simplified by constraining their motion with external features, for example against a table~\cite{yan2019self, zhu2018dual, nair2017combining}, with additional grippers~\cite{mayer2008system}, or pegs~\cite{saha2008motion}. Another common strategy involves limiting movements to long series of small deformations with pick and place actions~\cite{yan2019self, nair2017combining}. Thus, the dynamics of the system can be treated as quasistatic.
%

Furthermore, all works above rely on stable grasps of the cable, but do not exploit cable sliding. Zhu~\textit{et al}. routed a cable around pegs using one end-effector that was attached to the cable end and another fixed end-effector that would passively let a cable slip through in order to pull out a longer length of cable~\cite{zhu2019robotic}. This system, while allowing for sliding, loses the ability to sense and control the state of the cable at the sliding end, which can be in contact with any point of the cable. 

Jiang ~\textit{et al}. traces cables in a wire harness using a gripper with rollers in the jaws ~\cite{jiang2015robotized}. This gripper passively adjusts grip force using springs to accommodate different sized cables. They sense and control the force perpendicular to the translational motion along the cable in order to follow the cable. The cables in our work are considerably smaller and less rigid, so such forces would be difficult to sense. 

Furthermore, both of the specialized, passive end-effectors in the above two works have limited capabilities beyond sliding along the cable. In our work, the parallel jaw gripper used to follow a cable is also used to insert the cable into a headphone jack, demonstrating the potential of this hardware setup for additional tasks.

Another example of work involving sliding with rope from Yamakawa ~\textit{et al}. shows how our framework could potentially be extended for the knot-tying task ~\cite{yamakawa2007permutation}. To pass one end of the rope through a loop, they leverage tactile sensing to roll the two rope ends relative to each other in between the fingers. 

\section{Tasks}
\myparagraph{Cable Following} 
The goal of the cable following task is to use a robot gripper to grip the beginning of the cable with proper force and then control the gripper to follow the cable contour all the way to its tail end. The beginning end of the cable is initially firmly gripped by another fixed gripper during the cable following process. The moving gripper is allowed to regrasp the cable by bringing it back to the fixed gripper, resulting in two-hand coordination with one of the hands fixed. Several cables with different properties (shape, stiffness, surface roughness) are tested here for generalization. 

\myparagraph{Cable Insertion} 
The goal of the cable insertion task is to find the plug at the end of a cable and insert the plug into the socket. We show that this can be done by leveraging the ability to slide the fingers on the cable, and demonstrate it with a headphone cable with a cylindrical jack connector at its end.

\myparagraph{Robot System} In order to tackle this task, the following four hardware elements are necessary: 
\begin{itemize}
\item  tactile sensor to measure the grasped cable position and orientation in real time
\item tactile sensor to measure the amount of friction force during sliding in real time
\item fast reactive gripper to modulate the grasping force according to the measured friction
\item fast reactive robot arm to follow the measured cable orientation and keep the measured cable position in the center of the gripper. 
\end{itemize}

\section{Method}

\begin{figure}[t]
	\centering
	\includegraphics[width=\linewidth]{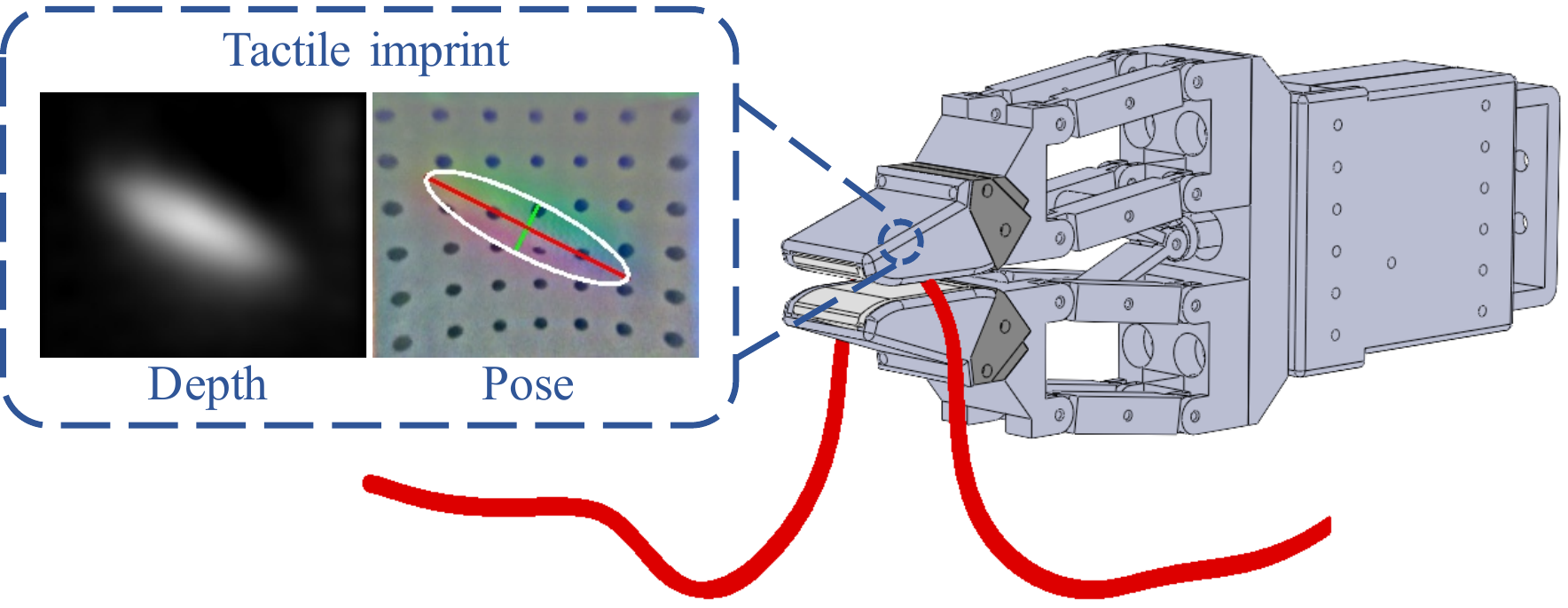}
	\vspace{-10pt}
	\caption{The design concept of the cable manipulation system.}
	\vspace{-10pt}
	\label{fig:gripper grasping a cable}
\end{figure}

The key idea of our method is to use tactile control to monitor and manipulate the position and forces on the cable between the fingers. The concept is illustrated in Fig.~\ref{fig:gripper grasping a cable}. We divide the tactile controller into two parts: 
\begin{itemize}
    \item[i.] \textbf{Cable Grip Control} so the cable alternates between firmly grasped and sliding smoothly,
    \item[ii.] \textbf{Cable Pose Control} so the cable remains centered and aligned with the fingers when pulling and sliding. 
\end{itemize}
In this section, we describe the implementation of the tactile controller by introducing the reactive gripper, the tactile perception system, the modeling of the cable, and the two controllers. 

\subsection{Hardware} 
\label{gripper}
\begin{figure}[t]
	\centering
	\includegraphics[width=0.9\linewidth]{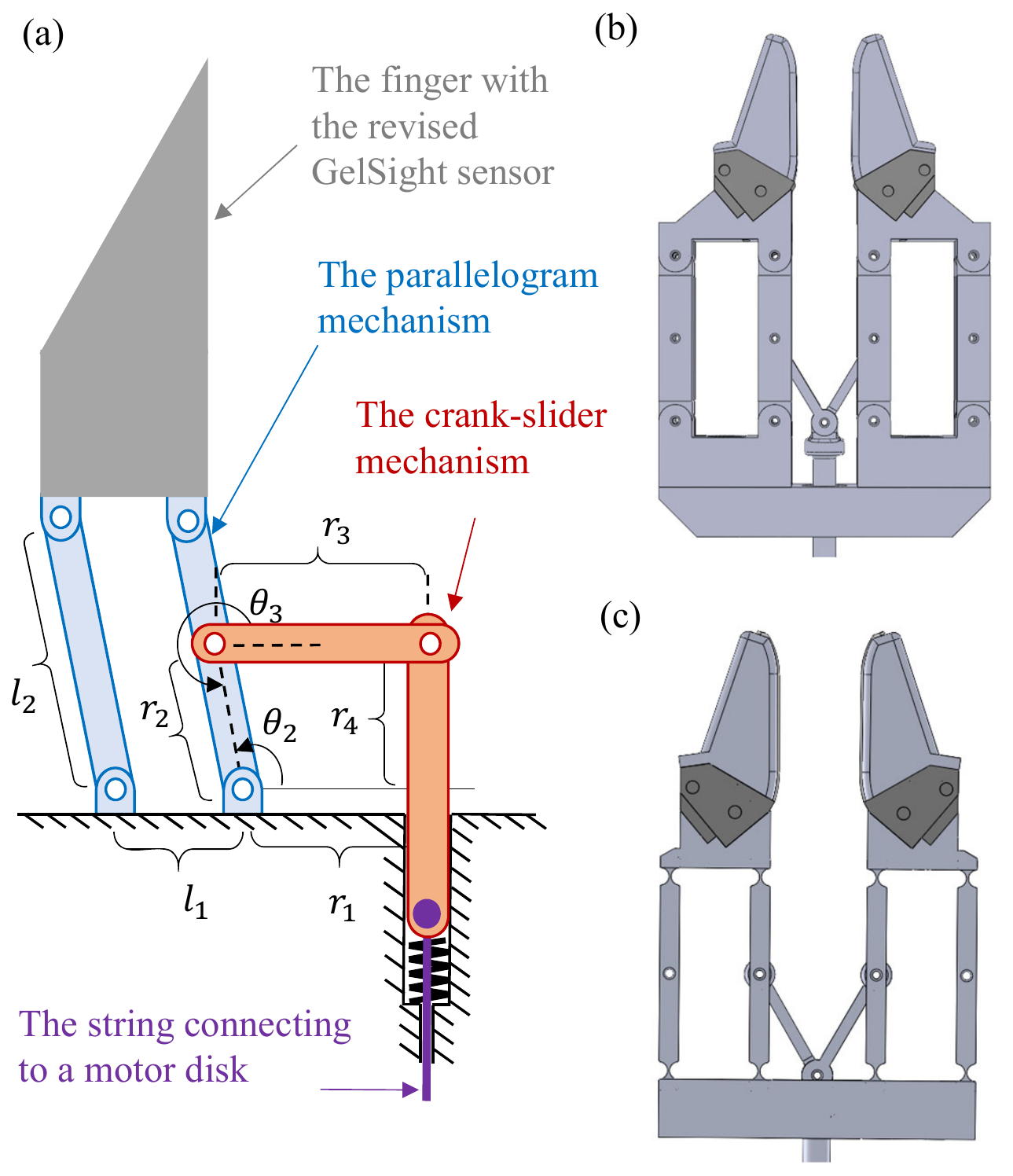}
	\vspace{-10pt}
	\caption{\textbf{Mechanism design.} (a) A motor drives the slider-crank mechanism via the slider-string-spring system, actuating the parallelogram mechanism via the crank linkage, and finally yielding the lateral displacement of the finger. (b) The rigid parallelogram mechanism includes 28 assembly parts. (c) The compliant parallel-guiding mechanism design replaces the rigid parallelogram mechanism reducing the assembly parts from 28 pieces to a single piece.}
	\vspace{-10pt}
	\label{fig:gripper design}
\end{figure}

Most commercialized robotic grippers do not offer sufficient bandwidth for real-time feedback control. To that end we design a parallel gripper with 2 fingers (with a revised GelSight sensor), a compliant parallel-guiding mechanisms, and slide-crank linkages  actuated by a servo motor (Dynamixel XM430-W210-T, Robotis) as shown in Fig.~\ref{fig:gripper design}. 

\myparagraph{Mechanism design}
Parallelogram mechanisms are widely used to yield lateral displacement and slider-crank mechanisms are broadly employed to actuate the parallelogram mechanism for parallel grippers. We use them to facilitate parallel grasping~\cite{modesitt1991micro}. To make a compact actuation mechanism, we use a tendon-driven system. 

One end of a string (tendon) is tied to a motor disk which is fixed on the servo motor installed in a motor case. The other end of the string is tied to the slider as shown in Fig.~\ref{fig:gripper design}a. We use a compression spring between the slider and the motor box with pre-tension forming a slider-string-spring system. 
One end of the crank is connected to the slider and the other is coupled with the rocker of the parallelogram mechanism. The finger is attached to the coupler of the parallelogram mechanism. The string drives the slider down, actuating the parallelogram mechanism via the crank linkage and producing the desired lateral displacement of the finger. Two fingers assembled symmetrically around the slider yields a parallel gripper. 

 \begin{figure*}[h]
	\centering
	\includegraphics[width=\linewidth]{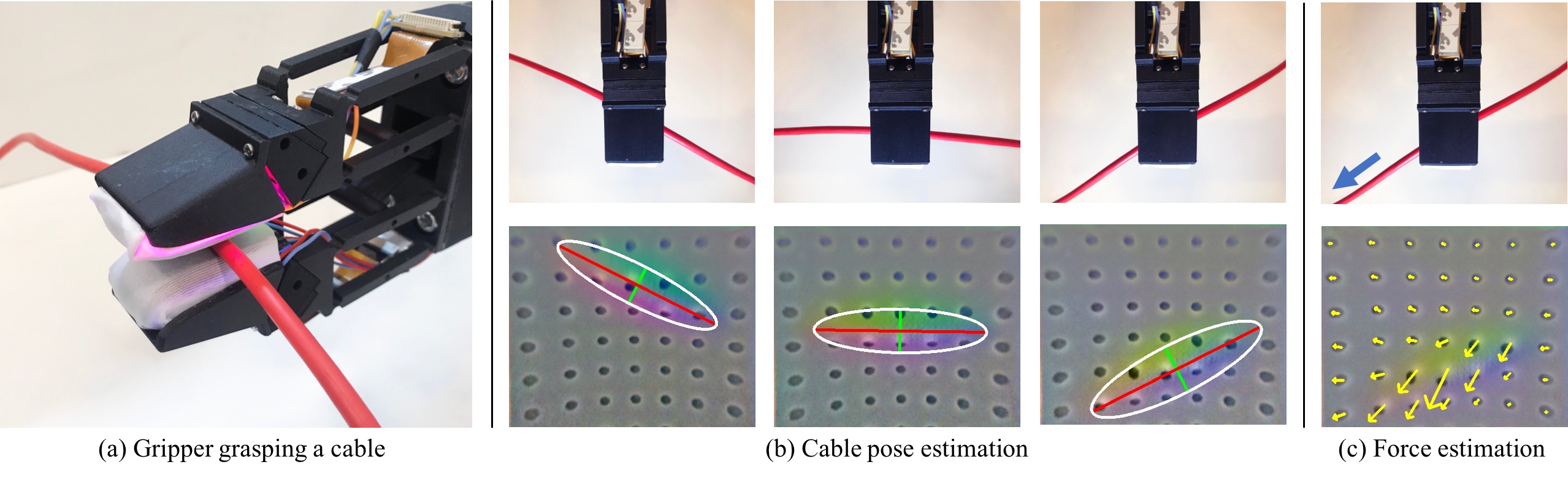}
    \vspace{-22pt}
	\caption{\textbf{Tactile perception.} (a) Gripper with GelSight sensors grasping a cable. (b) Top view of the gripper grasping different cable configurations and the corresponding cable pose estimations. The white ellipse shows the estimation of the contact region. The red and green lines show the first and second principal axes of the contact region, with lengths scaled by their eigenvalues. (c) Top view of pulled cable while the gripper registers marker displacements indicating the magnitude and direction of the frictional forces.}
    \vspace{-12pt}
	\label{fig:signal}
\end{figure*}

\myparagraph{Mechanism dimensions}
The design guidelines are as follows: 1) The max opening of the gripper is targeted at 100 mm, i.e., 50 mm displacement for each finger; 2) The parallelogram mechanism should fit the size of GelSight fingertips; 3) Reduce overall size of the gripper. With these constraints, we found a candidate dimension of the mechanisms as: 
$r_{1}$   =  15 mm; 
$r_{2}$   = 50 mm; 
$r_{3}$  = 30 mm; 	
$r_{4}$   = 20 mm; 
$l_{1}$	= 23.25 mm; 
$l_{2}$ = 	100 mm; 
$\theta_{2}^{i}$ = 127 $^{\circ}$; 
$\theta_{3}^{i}$ = 307 $^{\circ}$.  
Refer to Fig.~\ref{fig:gripper design} for the definition of all variables. Note that $\theta_{2}^{i}$ and $\theta_{3}^{i}$ are the initial values of $\theta_{2}$ and $\theta_{3}$.


\myparagraph{Compliant joint design}
Compliant mechanisms \cite{howell2001compliant} can produce exactly the same motion as those produced by rigid body mechanisms, but greatly reduce the part count and assembly time. We use compliant joints to simplify the parallelogram mechanism. The Pseudo-Rigid-Body (PRB) model bridges between compliant mechanisms and rigid body mechanisms \cite{howell2001compliant}. The PRB model is widely used to analyze kinematics and statics (forward method) of compliant mechanisms. One may develop a compliant mechanism based on the corresponding rigid body mechanism by applying an inverse analysis of the PRB model. In this study, we leverage the inverse method and convert the rigid parallelogram mechanism to an equivalent compliant parallel-guiding mechanism to reduce the assembly size. 

The rigid parallelogram mechanism in Fig.~\ref{fig:gripper design}b contains 28 pieces. We redesign the linkage with flexural joints (Fig.~\ref{fig:gripper design}c), reducing the 28 pieces of the rigid mechanism to a single part offering the same kinematic functionality.
%
%
The overall size of the final prototype has length $260$ mm, width $140$ mm, and thickness $85$ mm at the rest position. 

\subsection{Perception} 
\label{sec:perception}

Figure~\ref{fig:signal} illustrates the process to extract cable pose, cable force and  grasp quality from tactile images. 

\myparagraph{Cable pose estimation} First, we compute depth images from GelSight images with Fast Poisson Solver~\cite{gelsight_journal}. Then, we extract the contact region by thresholding the depth image. Finally, we use Principal Component Analysis (PCA) on the contact region to get the principal axis of the imprint of the cable on the sensor. 

\myparagraph{Cable friction force estimation} We use blob detection to locate the center of the black markers in the tactile images~\cite{gelslim_slip}. Then we use a matching algorithm to associate marker positions between frames, with a regularization term to maintain the smoothness of the marker displacement flow. We compute the mean of the marker displacement field (D), which is approximately proportional to the friction force.

\begin{figure}[b]
	\centering
	\includegraphics[width=1.0\linewidth]{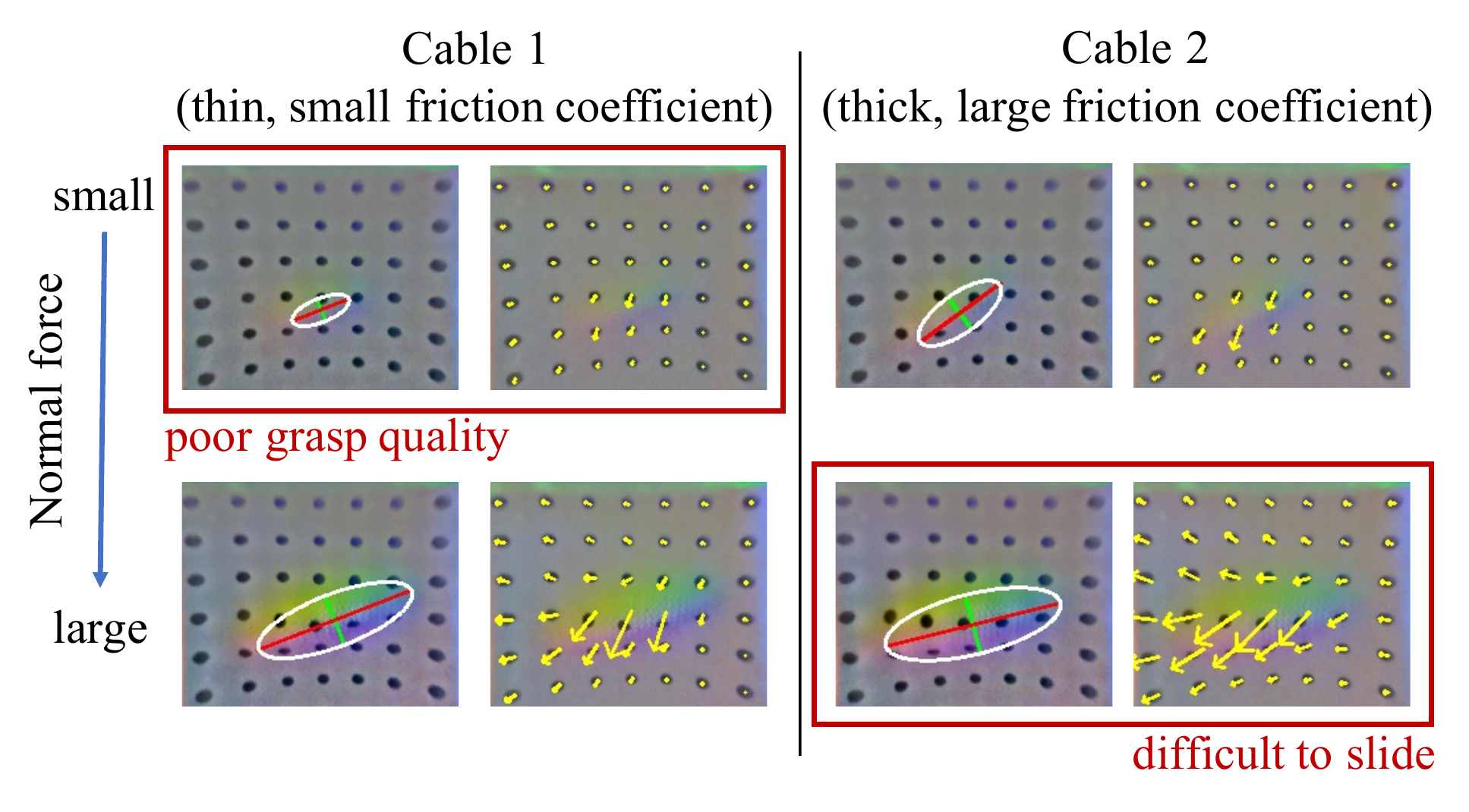}
	\vspace{-20pt}
	\caption{\textbf{Trade-off between tactile quality and sliding friction.} Larger gripping forces lead to higher-quality tactile imprints but difficult sliding. With the same normal force, the grasp quality and friction force varies among cables. The tactile-reactive control adjusts to different cables.}
	\label{fig:quality}
\end{figure}

\myparagraph{Cable grasp quality} In this task, we evaluate the grasp quality (S) based on whether the area of the contact region is larger than a certain area. A tactile imprint with poor quality (small contact region) will give noisy and uncertain pose estimation. By increasing the grasping force, as shown in Fig.~\ref{fig:quality}, we can increase the grasp quality.

\subsection{Control} 
\label{sec:control}

\myparagraph{Cable Grip Controller}
The goal of the grip controller is to modulate the grasping force such that 1) the friction force stays within a reasonable value for cable sliding (too small and the cable falls from the grip, too large and the cable gets stuck), and 2) the tactile signal quality is maintained. We employ a combination of a PD controller and a leaky integrator. The PD controller uses the mean value of the marker displacement (D) (correlating to the approximate friction force) as feedback and regulates it to a predefined target value ($D_{t}$). We use position control to modulate gripping force, and the PD controller for the reference velocity $u_{pd}$ is expressed as:
\begin{equation}
  \label{eq:1}
  \begin{gathered}
    u_{pd}[n] = K_pe[n] + K_d(e[n]-e[n-1])\\
    e[n] = D[n] - D_{t}[n],
  \end{gathered}
\end{equation}
where $K_p$ and $K_d$ are the coefficients for the proportional and derivative terms, and $D[n]$ is the measured mean value of the marker displacement. The leaky integrator raises $D_{t}$ of the PD controller if the signal quality (S) is poor as follows:
\begin{equation}
  \label{eq:2}
  \begin{gathered}
    D_t[n] = \alpha D_t[n-1] + (1-\alpha)(1-S),\\
    S = 
    \begin{cases}
    1  & \text{if good quality} \\
    0  & \text{if poor quality} 
    \end{cases}\\
  \end{gathered}
\end{equation}
where $\alpha$ is the leakage at each time step and S is the signal quality indicator. 

\begin{figure}[t]
	\centering
	\includegraphics[width=\linewidth]{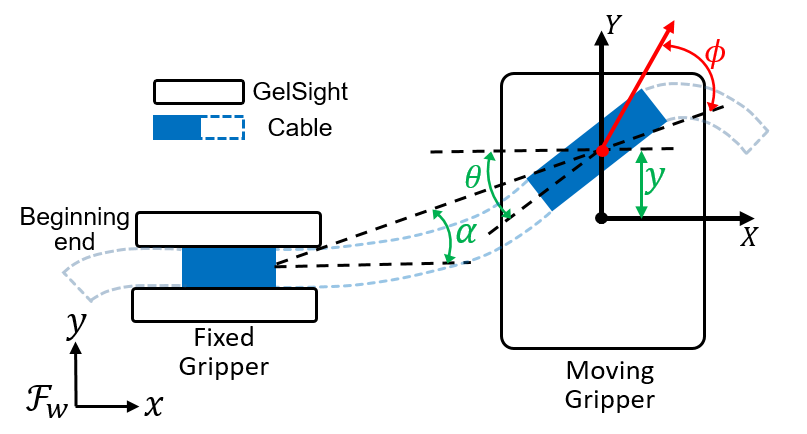}
	\vspace{-15pt}
	\caption{\textbf{Model cable-gripper dynamics.} Schematic diagram of the planar cable pulling modeling.}
	\vspace{-10pt}
	\label{fig:schematic}
\end{figure}

\myparagraph{Model of Cable-Gripper Dynamics}
We model the cable-gripper dynamics as a planar pulling problem. As shown in Fig.~\ref{fig:schematic}, the region of the cable in contact with the tactile sensor (blue rectangle on the right) is represented as a 2D object. We parameterize its position and orientation with respect to $X$ axis of the sensor frame with $y$ and $\theta$. 
We further define the angle $\alpha$ between the center of the cable on the moving gripper and the orientation of the fixed gripper (blue rectangle on the left).
%
These three parameters $[y \,\, \theta \,\, \alpha]$ define the state of the cable-gripper system.
We finally define the input of control on the system $\phi$ as the pulling direction relative to the angle $\alpha$ (labeled with red arrow). 

Since a deformable gel surface has complex friction dynamics, we use a data-driven method to fit a linear dynamic model rather than first principles. The state of the model is $\textbf{x}=[y \,\, \theta \,\, \alpha]^T$, the control input $\textbf{u} = [\phi]$, and the linear dynamic model: 
\begin{equation}
  \label{eq:3}
  \begin{gathered}
    \dot{\textbf{x}} = A \textbf{x} + B \textbf{u},
  \end{gathered}
\end{equation}
where $A$ and $B$ are the linear coefficients of the model.

To efficiently collect data, we use a simple proportional (P) controller as the base controller supplemented with uniform noise for the data collection process. The P controller controls the velocity of the robot TCP in the $y$ axis and we leave the velocity in the $x$ axis constant. The controller is expressed in Equation~\ref{eq:4}, where $K_{p}^v$ is the coefficient of the proportional term, and $N[n]$ is random noise sampled from a uniform distribution $[-0.01,0.01]$. The intuition for this baseline controller is that when the robot (sensor) moves to the $+\vec{y}$ direction, the cable gets pulled from the opposite direction $-\vec{y}$ and dragged back to the center of the gripper if it is initially located in the $+\vec{y}$ region.

\begin{equation}
  \label{eq:4}
  \begin{gathered}
    v_y[n] = K_{p}^vy[n] + N[n]
  \end{gathered}
\end{equation}

We collect approximately 2000 data points with a single cable by running several trajectories with different initial cable configurations. We use 80\% of the data for linear regression of matrices $A$ and $B$ and validate the result with the rest. 

\myparagraph{Cable Pose Controller} 
The goal of the cable pose controller is to maintain the cable position in the center of the GelSight sensor ($y^* = 0$), the orientation of the cable to be parallel to the $X$ axis ($\theta^* = 0$) and the inclination of the pulled cable also parallel to the $X$ axis ($\alpha^* = 0$).
The nominal trajectory of the cable pose controller ($\textbf{x[n]}^*$, $\textbf{u[n]}^*$) is then constant and equal to ($[0\,\,0\,\,0]^T$, $[0]$), that is, regulating around zero.

We formulate an LQR controller with the $A$ and $B$ matrices from the linear model and solve for the optimal feedback gain $K$, which in turn gives us the optimal control input $\bar{\textbf{u}}[n] = -K\textbf{x}[n]$, where $\textbf{x}=[y \,\, \theta \,\, \alpha]^T$ as shown in fig~\ref{fig:schematic}.
The parameters of the LQR controller we use are $\textbf{Q} = [1,1,0.1]$ and $\textbf{R} = [0.1]$, since regulating $y$ and $\theta$ (making sure the cable does not fall) is more important than regulating $\alpha$ (maintain the cable straight).

\section{Experiment}

\subsection{Experimental Setup}  
\begin{figure}[h]
	\centering
	\includegraphics[width=\linewidth]{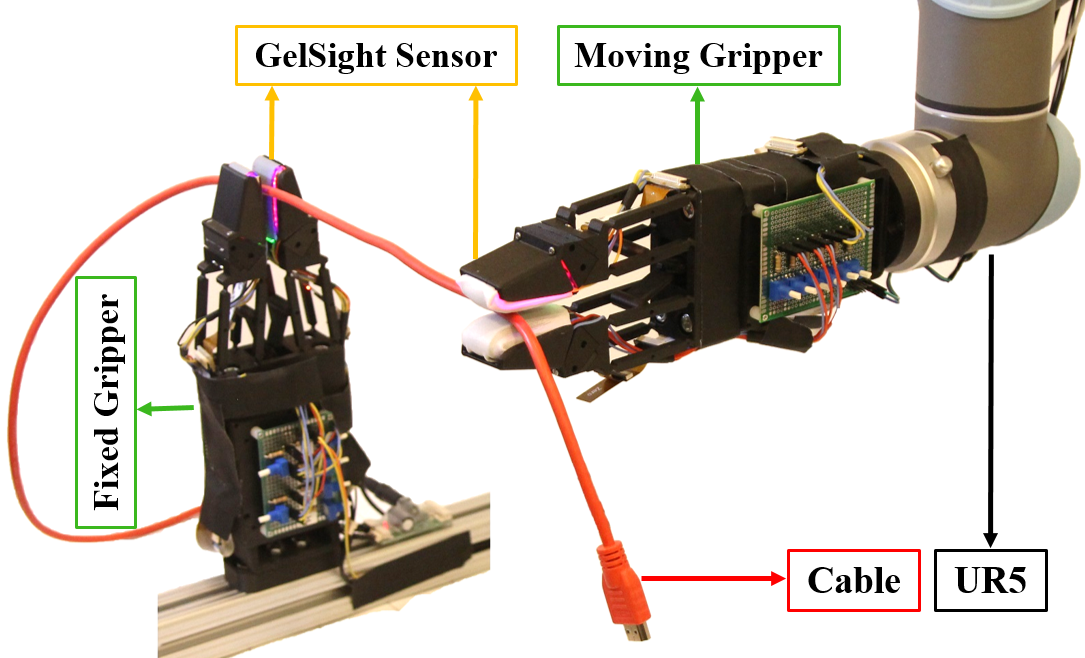}
    \vspace{-15pt}
	\caption{\textbf{Experimental setup.} UR5 robot arm and two reactive grippers with GelSight sensors.}
	\vspace{-10pt}
	\label{fig:setup}
\end{figure}

The experimental setup in Fig.~\ref{fig:setup} includes a 6-DOF UR5 robot arm, two reactive grippers (as described in Section~\ref{gripper}) and two pairs of revised fingertip GelSight sensors attached to the gripper fingers. One of the grippers is fixed on the table and another one is attached to the robot. The control loop frequencies of the UR5 and the gripper are 125 Hz and 60 Hz, respectively. 

We use five different cables/ropes to test the controllers (Fig.~\ref{fig:Results} bottom): Thin USB cable with nylon surface; thick HDMI cable with rubber surface; thick nylon rope; thin nylon rope; and thin USB cable with rubber surface.


\subsection{Cable following experiments} 
\myparagraph{Experimental process} The beginning end of the cable is initially grasped firmly with the fixed gripper secured at a known position. The moving gripper picks up the cable and follows it along its length until reaching its tail end. During that process, the grasping force is modulated with the cable grip controller and the pose of the cable is regulated with the cable pose controller. The moving gripper can regrasp the cable by feeding the holding part to the fixed gripper if it feels it is going to loose control of the cable, or the robot reaches the workspace bounds. Within a regrasp, the robot adjusts the position of the moving gripper according to the position of the cable detected by the fixed gripper.

\myparagraph{Metrics} we use three metrics to evaluate performance:
\begin{itemize}
\item Ratio of cable followed vs. total cable length 
\item Distance traveled per regrasp, normalized by the maximum workspace of the moving gripper.
\item Velocity, normalized by max velocity in the $x$ direction.
\end{itemize}
Note that all these metrics have a max and ideal value of 1.

\myparagraph{Controller comparison} We compare the proposed LQR controller with three baselines: 1) purely moving the robot to the $x$ direction without any feedback (open-loop controller), 2) open-loop controller with emergency regrasps before losing the cable, 3) Proportional (P) controller we use to collect data. Since the initial configuration of the cable affects the result dramatically, we try to keep the configuration as similar as possible for the control experiments and average the results for 5 trials of each experiment. 

\myparagraph{Generalization} We conduct control experiments with the LQR robot controller + PD gripper controller to test the performance across 1) different velocities: 0.025, 0.045, and 0.065 m/s; and 2) 5 different cables. Similarly, we also conduct 5 trials for each experiment and average the results.

\subsection{Cable following and insertion experiment} 
An illustrative application of the cable following skill is to find a connector at the end of a cable to insert it. This is a robust strategy to find the connector end of a cable when it is not directly accessible or under position uncertainties. Here we conduct an experiment on a headphone cable. The relative position of the hole where to insert the connector is calibrated. The cable following process is identical to what we illustrated in the previous section. We detect the plug (thick) based on its geometry difference compared to the cable (thin) using GelSight sensor. To estimate the pose of the plug before insertion, we use the same tactile estimation method as used to estimate the cable pose  during cable following.





\section{Experimental Results}
\label{sec:results}

In this section, we detail the results of the cable following experiments with different robot controllers, different velocities and different cables, according to the evaluation metrics. See Fig.~\ref{fig:Results} for a summary. We also show the results of the cable following and insertion experiment (Fig.~\ref{fig:insertion}). 

\begin{figure}[t]
	\centering

	\includegraphics[width=1.0\linewidth]{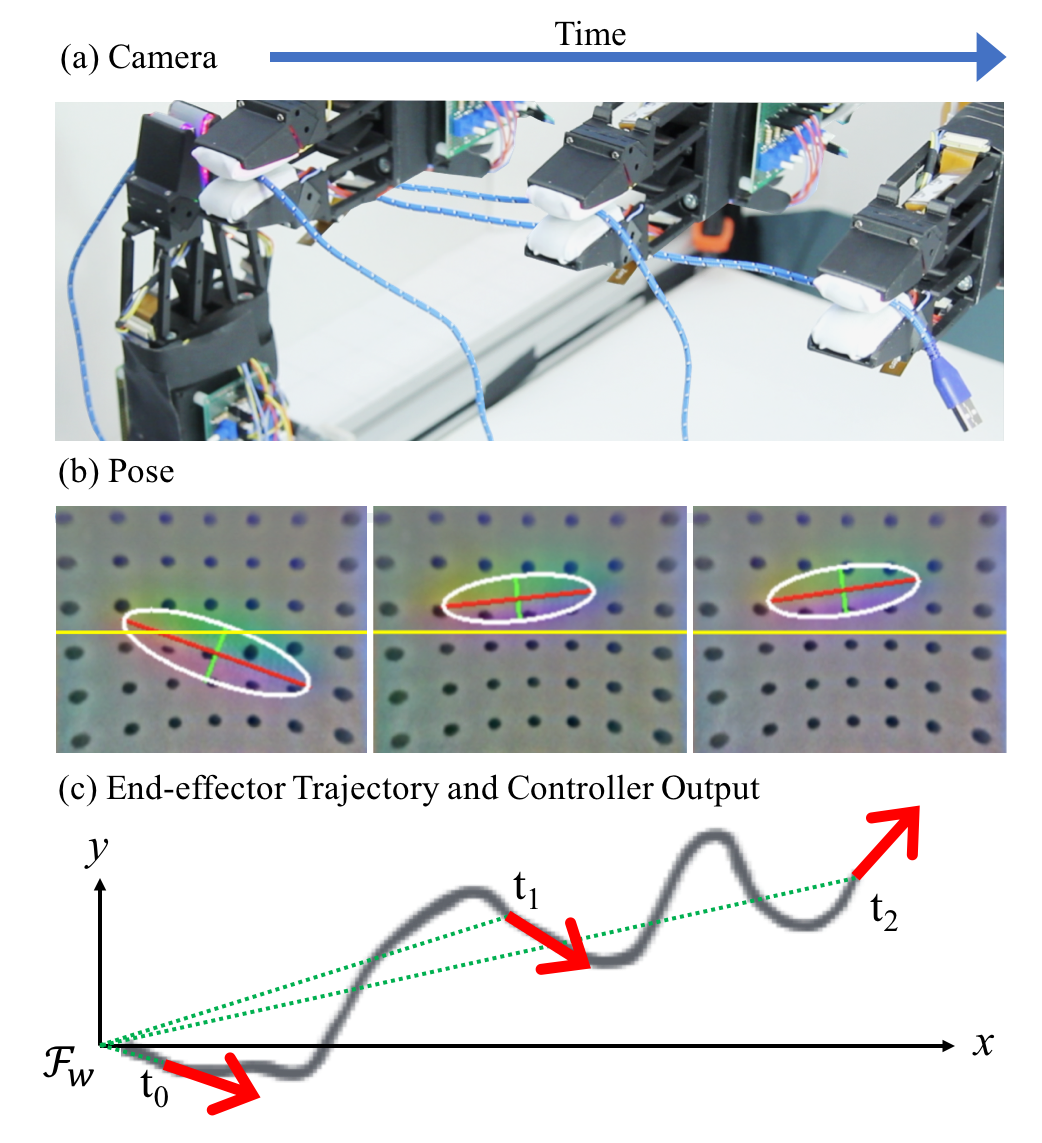}
    \vspace{-13pt}
	\caption{\textbf{Cable following experiment.} For three instances in time, (a) camera view; (b) pose estimation from tactile imprints,  where the yellow line in the center indicates the desired in-hand pose alignment; (c) top view of the trajectory of the end-effector and velocity output of the LQR controller, shown in red. The green dotted line illustrates $\alpha$. The controller keeps adjusting the cable state in real-time by changing the moving direction to achieve the desired pulling angle.} 
	\vspace{-10pt}
	\label{fig:control}
\end{figure}
\subsection{Controller evaluation}
We compare four different robot controllers: open-loop, open-loop with emergency regrasps, P controller, and LQR controller. The top row in Fig.~\ref{fig:Results} shows that the open-loop controller only follows 36\% of the total length. The gripper loses the cable easily when it curves. The simple addition of emergency regrasps is sufficient for the open loop controller to finish the task. This indicates that a timely detection of when the cable is about to fall from the gripper is important for this task.
This controller, however, still requires many regrasps and is slower than the P and the LQR controllers. The results show that the LQR controller uses the least number of regrasps compared to other controllers. The LQR controller does not show much experimental improvement in the velocity metric, in part because the robot travels more trying to correct for cable deviations. 

Figure~\ref{fig:control} shows snapshots of the experimental process using the LQR controller. Note that this controller always tries to move the gripper back to the center of the trajectory once the cable is within the nominal configuration since $\alpha$ (the angle between the center of the cable in hand to the beginning end) is also part of the state. This can be observed from the middle time instance of Fig.~\ref{fig:control}, where the cable is straight and close to the middle of the GelSight sensor, but $\alpha$ is large (the angle between the x axis and the green line to $t_1$ in the bottom of Fig.~\ref{fig:control}). The output of the controller shows the direction to the center of the trajectory. The pose in the last time instance is similar, but because $\alpha$ is smaller, the controller outputs a direction that will correct the cable pose. This feature is an advantage of the LQR controller over the P controller.





\subsection{Generalization to different velocities}
\begin{figure}
	\centering
	\includegraphics[width=1.0\linewidth]{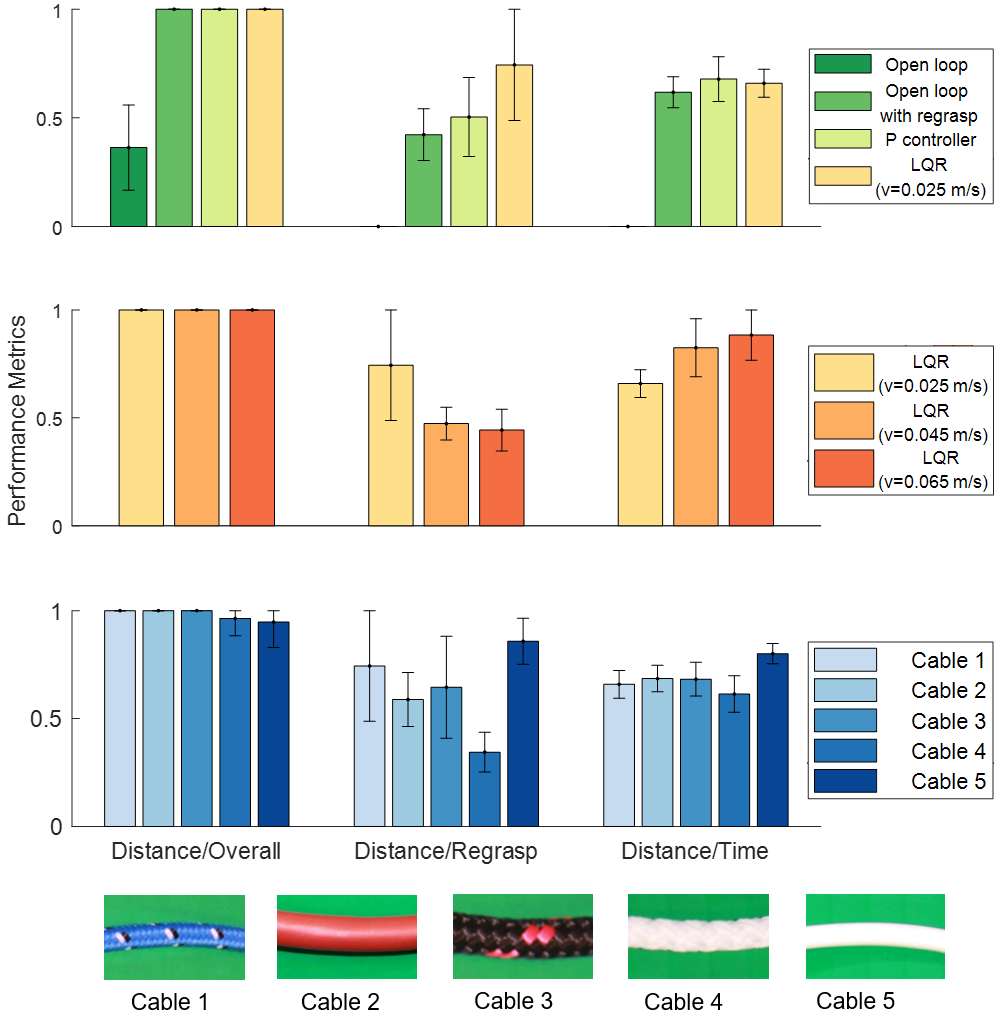}
	\vspace{-1pt}
	\caption{\textbf{Experimental results.} Different robot controllers (top), different following velocities (middle), different cables (bottom). For visualization, the three metrics are normalized to [0, 1] by dividing 100\%, 0.45 m, and 0.02 m/s respectively (max 1, ideal 1).}
	\vspace{-15pt}
	\label{fig:Results}
\end{figure}

The model of the cable-gripper dynamics is fit with data collected at a velocity of 0.025 m/s. We also test the LQR controller at 0.045 and 0.065 m/s. The results in the second row of Fig.~\ref{fig:Results} show that the performance does not degrade, except requiring more regrasps per unit of distance travelled. This is because, going faster, the controller has less time to react to sudden pose changes and, therefore, tends to trigger regrasps more. Although the number of regrasps increases with larger velocity, the total time is still shorter due to the faster velocity.

\begin{figure*}[t]
	\centering

	\includegraphics[width=1.0\linewidth]{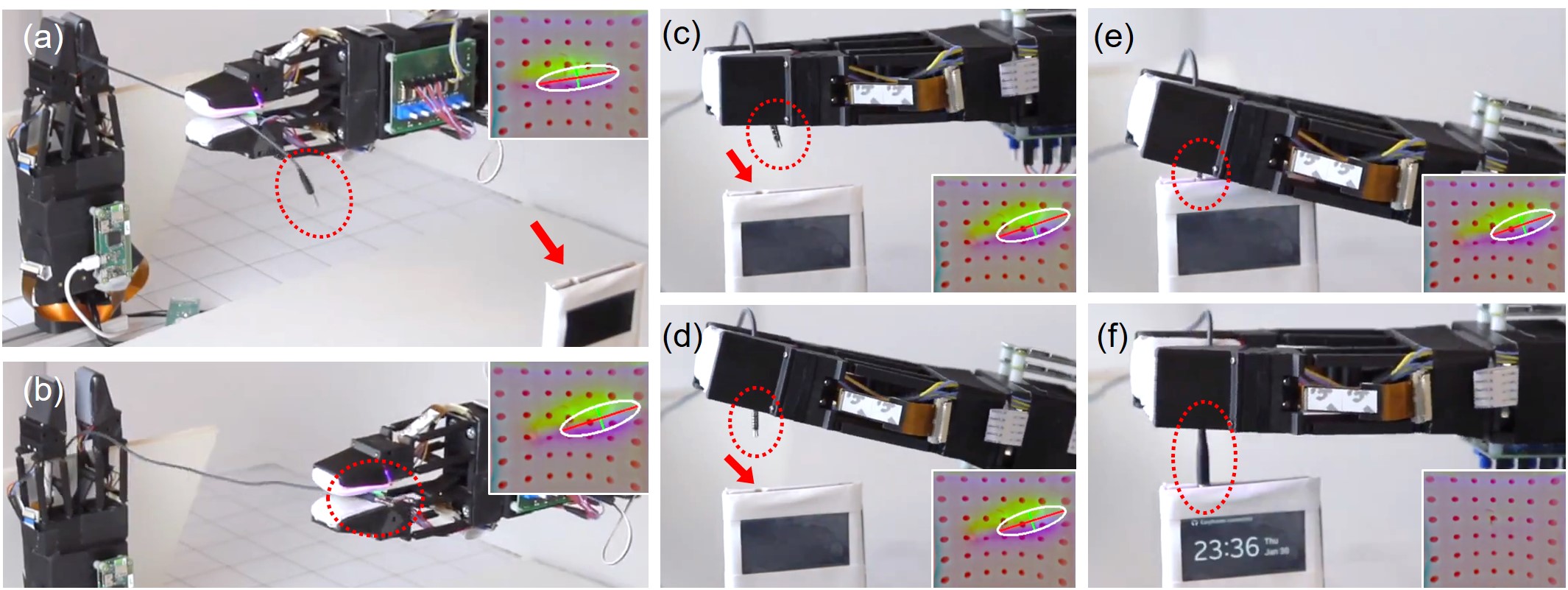}
    \vspace{-13pt}
	\caption{\textbf{Headphone cable following and insertion process.} (a)(b) cable following to the plug end, (c) plug on top of the hole with pose mismatch (d) plug pose adjusted and aligning with the hole, (e)(f) cable plugged into the headphone jack on the phone. The plug is labeled with red circle and the headphone jack is labeled with red arrow. } 
	\vspace{-10pt}
	\label{fig:insertion}
\end{figure*}

\subsection{Generalization to different cables}
We also test the system with the LQR controller on 5 different cables, each with different physical properties (diameters, materials, stiffness). In experiments, the system generalizes well and can follow 98.2\% of the total length of the cables. 

Comparing the performance on the different cables shows that cable 4 (thin and light nylon rope) requires the most regrasps. It is difficult to adjust in-hand pose since it is very light and the un-followed part of the cable tends to move with the gripper. The cable with best performance is cable 5 (thin and stiff rubber USB cable), which is stiff and locally straight at most of the time.


\subsection{Cable following and insertion}
The process to grasp, follow, and insert the headphone cable is illustrated in Fig.~\ref{fig:insertion}. Parts (a) and (b) show the robot following the cable all the way to the plug and identifying the moment it reaches the plug. After the plug is detected, the fixed gripper opens and the robot moves the plug over the headphone jack on the phone shown in Fig.~\ref{fig:insertion}(c). 

After cable sliding, the gripper uses the tactile feedback from a GelSight sensor to localize the plug and align it with the hole, as shown in Fig.~\ref{fig:insertion}(d). Afterwards, the robot moves down to insert the cable into the headphone jack in Fig.~\ref{fig:insertion}(e)-(f). We repeat the insertion only experiment (plug directly fed to the gripper by human with random pose) for 20 times and can insert the headphone plug with an 85\% success rate. See the attached supplementary video for better visualization of combined experiments of cable following and insertion.

\section{Conclusions and Discussion}
In this paper, we present a perception and control framework to tackle the cable following task. 
We show that the tight integration of tactile feedback, gripping control, and robot motion, jointly with a sensible decomposition of the control requirements is key to turning the--a priori--complex task of manipulating a highly deformable object with uncontrolled variations in friction and shape, into an achievable task. 
The main contributions of the work are:

\begin{itemize}
\item \textbf{Tactile Perception} Applying vision-based tactile sensor, like GelSight, to cable manipulation tasks. It provides rich but easy-to-interpret tactile imprints for tracking the cable pose and force in real-time. These local pose and force information about the cables are otherwise difficult to be captured from vision system during continuous manipulation because they are usually occluded, expensive to interpret, and not accurate enough.



\item \textbf{Tactile Gripper} The design of reactive gripper uses compliant joints, making it easy to fabricate, and protecting the tactile sensor from unexpected collision. The gripper modulates grasping force at 60 Hz which enables tasks that need fast response to tactile feedback like cable manipulation. 



\item \textbf{Tactile Control} We divide the control of the interaction between cable and gripper into two controllers: 1) Cable Grip Controller, a PD controller and leaky integrator that maintain an adequate friction level between gripper and cable to allow smooth sliding; and 2) Cable Pose Controller, an LQR controller based on a linearized model of the gripper-cable interaction dynamics, that maintains the cable centered and aligned with the fingers. 
\end{itemize}

%

The successful implementation of the tactile perception and model-based controller in the cable following task, and its generalization to different cables and to different following velocities, demonstrates that it is possible to use simple models and controllers to manipulate deformable objects.
The illustrative demonstration of picking and finding the end of a headphone cable for insertion provides a example of how the proposed framework can play a role in practical cable-related manipulation tasks.

There are several aspects of the system that can be improved: 1) The frequency of the tactile signal (30Hz) and the control loop of the gripper (60Hz) can be potentially raised to 90Hz and 200Hz, respectively; 2) We observe that it is difficult to pull the cable back when it reaches the edge of the finger, because of the convex surface of the GelSight sensor. The finger-sensor shape could be better optimized to improve performance; 3) It would be interesting to explore other models and controllers. Model-based reinforcement learning with a more complex function approximator could be a good fit for this task to handle with more accuracy the cable-gripper dynamics. 
%
The perception and control frameworks proposed here might enable addressing more complex robotic tasks. 



\bibliographystyle{IEEEtran}
\bibliography{ref}

\begin{thebibliography}{10}
\providecommand{\url}[1]{#1}
\csname url@rmstyle\endcsname
\providecommand{\newblock}{\relax}
\providecommand{\bibinfo}[2]{#2}
\providecommand\BIBentrySTDinterwordspacing{\spaceskip=0pt\relax}
\providecommand\BIBentryALTinterwordstretchfactor{4}
\providecommand\BIBentryALTinterwordspacing{\spaceskip=\fontdimen2\font plus
\BIBentryALTinterwordstretchfactor\fontdimen3\font minus
  \fontdimen4\font\relax}
\providecommand\BIBforeignlanguage[2]{{%
\expandafter\ifx\csname l@#1\endcsname\relax
\typeout{** WARNING: IEEEtran.bst: No hyphenation pattern has been}%
\typeout{** loaded for the language `#1'. Using the pattern for}%
\typeout{** the default language instead.}%
\else
\language=\csname l@#1\endcsname
\fi
#2}}

\bibitem{yan2019self}
M.~Yan, Y.~Zhu, N.~Jin, and J.~Bohg,
  ``\href{https://arxiv.org/pdf/1911.06283.pdf} {Self-Supervised Learning of
  State Estimation for Manipulating Deformable Linear Objects},'' \emph{arXiv
  preprint arXiv:1911.06283}, 2019.

\bibitem{zhu2018dual}
J.~Zhu, B.~Navarro, P.~Fraisse, A.~Crosnier, and A.~Cherubini,
  ``\href{https://ieeexplore.ieee.org/document/8593780} {Dual-arm robotic
  manipulation of flexible cables},'' in \emph{2018 IEEE/RSJ International
  Conference on Intelligent Robots and Systems (IROS)}.\hskip 1em plus 0.5em
  minus 0.4em\relax IEEE, 2018, pp. 479--484.

\bibitem{nair2017combining}
A.~Nair, D.~Chen, P.~Agrawal, P.~Isola, P.~Abbeel, J.~Malik, and S.~Levine,
  ``\href{https://arxiv.org/abs/1703.02018} {Combining self-supervised learning
  and imitation for vision-based rope manipulation},'' in \emph{2017 IEEE
  International Conference on Robotics and Automation (ICRA)}.\hskip 1em plus
  0.5em minus 0.4em\relax IEEE, 2017, pp. 2146--2153.

\bibitem{gelsight_journal}
W.~Yuan, S.~Dong, and E.~Adelson,
  ``\href{https://www.mdpi.com/1424-8220/17/12/2762/htm}{Gelsight:
  High-resolution robot tactile sensors for estimating geometry and force},''
  \emph{Sensors}, vol.~17, no.~12, p. 2762, 2017.

\bibitem{lange1998viscontour}
F.~Lange, P.~Wunsch, and G.~Hirzinger,
  ``\href{https://ieeexplore.ieee.org/stamp/stamp.jsp?arnumber=680743}{Predictive
  Vision Based Control of High Speed Industrial Robot Paths},'' in \emph{1998
  IEEE/RSJ International Conference on Robotics and Automation (ICRA)}.\hskip
  1em plus 0.5em minus 0.4em\relax IEEE, 1998.

\bibitem{chen1995edge}
N.~Chen, H.~Zhang, and R.~Rink,
  ``\href{https://ieeexplore.ieee.org/document/526143/}{Edge tracking using
  tactile servo},'' in \emph{Proceedings 1995 IEEE/RSJ International Conference
  on Intelligent Robots and Systems. Human Robot Interaction and Cooperative
  Robots}, vol.~2.\hskip 1em plus 0.5em minus 0.4em\relax IEEE, 1995, pp.
  84--89.

\bibitem{tactip}
B.~Ward-Cherrier, N.~Pestell, L.~Cramphorn, B.~Winstone, M.~E. Giannaccini,
  J.~Rossiter, and N.~F. Lepora,
  ``\href{https://www.liebertpub.com/doi/pdfplus/10.1089/soro.2017.0052}{The
  tactip family: Soft optical tactile sensors with 3d-printed biomimetic
  morphologies},'' \emph{Soft robotics}, vol.~5, no.~2, pp. 216--227, 2018.

\bibitem{hellman2017functional}
R.~B. Hellman, C.~Tekin, M.~van~der Schaar, and V.~J. Santos,
  ``\href{https://ieeexplore.ieee.org/stamp/stamp.jsp?arnumber=8039205}{Functional
  contour-following via haptic perception and reinforcement learning},''
  \emph{IEEE transactions on haptics}, vol.~11, no.~1, pp. 61--72, 2017.

\bibitem{hopcroft1991case}
J.~E. Hopcroft, J.~K. Kearney, and D.~B. Krafft,
  ``\href{https://journals.sagepub.com/doi/pdf/10.1177/027836499101000105}{A
  case study of flexible object manipulation},'' \emph{The International
  Journal of Robotics Research}, vol.~10, no.~1, pp. 41--50, 1991.

\bibitem{morita2003knot}
T.~Morita, J.~Takamatsu, K.~Ogawara, H.~Kimura, and K.~Ikeuchi,
  ``\href{https://ieeexplore.ieee.org/stamp/stamp.jsp?arnumber=1242193}{Knot
  planning from observation},'' in \emph{2003 IEEE International Conference on
  Robotics and Automation (Cat. No. 03CH37422)}, vol.~3.\hskip 1em plus 0.5em
  minus 0.4em\relax IEEE, 2003, pp. 3887--3892.

\bibitem{saha2008motion}
M.~Saha, P.~Isto, and J.-C. Latombe,
  ``\href{http://ai.stanford.edu/~latombe/papers/iser-06/paper.pdf}{Motion
  planning for robotic manipulation of deformable linear objects},''
  \emph{Experimental Robotics}, p. 23–32, 2008.

\bibitem{mayer2008system}
H.~Mayer, F.~Gomez, D.~Wierstra, I.~Nagy, A.~Knoll, and J.~Schmidhuber,
  ``\href{http://people.idsia.ch/~tino/papers/mayer.iros06.pdf}{A system for
  robotic heart surgery that learns to tie knots using recurrent neural
  networks},'' \emph{Advanced Robotics}, vol.~22, no. 13-14, pp. 1521--1537,
  2008.

\bibitem{zhu2019robotic}
J.~Zhu, B.~Navarro, R.~Passama, P.~Fraisse, A.~Crosnier, and A.~Cherubini,
  ``\href{https://ieeexplore.ieee.org/stamp/stamp.jsp?arnumber=8851170}{Robotic
  Manipulation Planning for Shaping Deformable Linear Objects With
  Environmental Contacts},'' \emph{IEEE Robotics and Automation Letters},
  vol.~5, no.~1, pp. 16--23, 2019.

\bibitem{jiang2015robotized}
X.~Jiang, Y.~Nagaoka, K.~Ishii, S.~Abiko, T.~Tsujita, and M.~Uchiyama,
  ``\href{https://www.sciencedirect.com/science/article/pii/S0736584514001069}
  {Robotized recognition of a wire harness utilizing tracing operation},''
  \emph{Robotics and Computer-Integrated Manufacturing}, vol.~34, pp. 52--61,
  2015.

\bibitem{yamakawa2007permutation}
Y.~Yamakawa, A.~Namiki, M.~Ishikawa, and M.~Shimojo,
  ``\href{https://ieeexplore.ieee.org/abstract/document/4399379}{One-handed
  knotting of a flexible rope with a high-speed multifingered hand having
  tactile sensors},'' in \emph{2007 IEEE/RSJ International Conference on
  Intelligent Robots and Systems (IROS)}.\hskip 1em plus 0.5em minus
  0.4em\relax IEEE, 2007.

\bibitem{modesitt1991micro}
D.~B. Modesitt, ``\href{https://patents.google.com/patent/EP0441060A2/da}
  {Micro-gripper assembly},'' Sept.~10 1991, uS Patent 5,046,773.

\bibitem{howell2001compliant}
L.~L. Howell,
  \emph{\href{https://link.springer.com/referenceworkentry/10.1007\%2F978-94-017-9780-1_302}
  {Compliant mechanisms}}.\hskip 1em plus 0.5em minus 0.4em\relax John Wiley \&
  Sons, 2001.

\bibitem{gelslim_slip}
S.~Dong, D.~Ma, E.~Donlon, and A.~Rodriguez,
  ``\href{https://arxiv.org/abs/1810.13381}{Maintaining Grasps within Slipping
  Bound by Monitoring Incipient Slip},'' in \emph{IEEE ICRA}, 2018.

\end{thebibliography}

\end{document}